\documentclass[journal]{IEEEtran}
\usepackage{graphicx, subfig}
\usepackage{diagbox}
\usepackage{amsmath}
\usepackage{amssymb}
\usepackage{multirow}
\usepackage{lineno,hyperref}
\usepackage[ruled, vlined]{algorithm2e}

\newcommand\bs[1]{\mathbf{#1}}
\ifCLASSINFOpdf
\else
\fi
\begin{document}
%
% paper title
% Titles are generally capitalized except for words such as a, an, and, as,
% at, but, by, for, in, nor, of, on, or, the, to and up, which are usually
% not capitalized unless they are the first or last word of the title.
% Linebreaks \\ can be used within to get better formatting as desired.
% Do not put math or special symbols in the title.
\title{Semantic Enhanced Knowledge Graph for Large-Scale Zero-Shot Learning}
%
%
% author names and IEEE memberships
% note positions of commas and nonbreaking spaces ( ~ ) LaTeX will not break
% a structure at a ~ so this keeps an author's name from being broken across
% two lines.
% use \thanks{} to gain access to the first footnote area
% a separate \thanks must be used for each paragraph as LaTeX2e's \thanks
% was not built to handle multiple paragraphs
%

\author{Jiwei~Wei, Yang~Yang*, Zeyu Ma, Jingjing~Li, Xing~Xu, Heng~Tao~Shen% <-this % stops a space
\thanks{*Corresponding Author: Yang Yang.}
\thanks{J. Wei, Y. Yang, Z. Ma, J. Li, X. Xu and H. T. Shen are with the Center for Future Media, University of Electronic Science and Technology of
China, Chengdu 611731, China, and also with the School of Computer
Science and Engineering, University of Electronic Science and Technology
of China, Chengdu 611731, China (e-mail: mathematic6@gmail.com;
dlyyang@gmail.com; cnzeyuma@163.com; xing.xu@uestc.edu.cn; lijin117@yeah.net; shenhengtao@hotmail.com). 
}% <-this % stops a space

%School of Computer Science and Engineering,
%University of Electronic Science and Technology of China.  Chengdu, Sichuan, China, 611731}% <-this % stops a space

%\thanks{Manuscript received April 19, 2005; revised August 26, 2015.}
}

% note the % following the last \IEEEmembership and also \thanks - 
% these prevent an unwanted space from occurring between the last author name
% and the end of the author line. i.e., if you had this:
% 
% \author{....lastname \thanks{...} \thanks{...} }
%                     ^------------^------------^----Do not want these spaces!
%
% a space would be appended to the last name and could cause every name on that
% line to be shifted left slightly. This is one of those "LaTeX things". For
% instance, "\textbf{A} \textbf{B}" will typeset as "A B" not "AB". To get
% "AB" then you have to do: "\textbf{A}\textbf{B}"
% \thanks is no different in this regard, so shield the last } of each \thanks
% that ends a line with a % and do not let a space in before the next \thanks.
% Spaces after \IEEEmembership other than the last one are OK (and needed) as
% you are supposed to have spaces between the names. For what it is worth,
% this is a minor point as most people would not even notice if the said evil
% space somehow managed to creep in.

% The paper headers
\markboth{Journal of \LaTeX\ Class Files,~Vol.~14, No.~8, August~2015}%
{Shell \MakeLowercase{\textit{et al.}}: Bare Demo of IEEEtran.cls for IEEE Journals}
% The only time the second header will appear is for the odd numbered pages
% after the title page when using the twoside option.
% 
% *** Note that you probably will NOT want to include the author's ***
% *** name in the headers of peer review papers.                   ***
% You can use \ifCLASSOPTIONpeerreview for conditional compilation here if
% you desire.

% If you want to put a publisher's ID mark on the page you can do it like
% this:
%\IEEEpubid{0000--0000/00\$00.00~\copyright~2015 IEEE}
% Remember, if you use this you must call \IEEEpubidadjcol in the second
% column for its text to clear the IEEEpubid mark.

% use for special paper notices
%\IEEEspecialpapernotice{(Invited Paper)}

% make the title area
\maketitle

% As a general rule, do not put math, special symbols or citations
% in the abstract or keywords.
\begin{abstract}
Zero-Shot Learning has been a highlighted research topic in both vision and language areas. Recently, most existing methods adopt structured knowledge information to model explicit correlations among categories and use deep graph convolutional network to propagate information between different categories. However, it is difficult to add new categories to existing structured knowledge graph, and deep graph convolutional network suffers from over-smoothing problem. In this paper, we provide a new semantic enhanced knowledge graph that contains both expert knowledge and categories semantic correlation.
Our semantic enhanced knowledge graph can further enhance the correlations among categories and make it easy to absorb new categories. To propagate information on the knowledge graph, we propose a novel Residual Graph Convolutional Network (ResGCN), which can effectively alleviate the problem of over-smoothing. Experiments conducted on the widely used large-scale ImageNet-21K dataset and AWA2 dataset show the effectiveness of our method, and establish a new state-of-the-art on zero-shot learning. Moreover, our results on the large-scale ImageNet-21K with various feature extraction networks show that our method has better generalization and robustness.
\end{abstract}

% Note that keywords are not normally used for peerreview papers.

% For peer review papers, you can put extra information on the cover
% page as needed:
% \ifCLASSOPTIONpeerreview
% \begin{center} \bfseries EDICS Category: 3-BBND \end{center}
% \fi
%
% For peerreview papers, this IEEEtran command inserts a page break and
% creates the second title. It will be ignored for other modes.
\IEEEpeerreviewmaketitle

%\linenumbers

%%%%%%%%% BODY TEXT
\section{Introduction}
\label{sec:intro}
\IEEEPARstart{R}{ecently}, with the fast development of deep learning techniques \cite{he2016deep, huang2017densely, wei2020universal, 9454290,wei2021meta,wei2022semantic} and large-scale high-quality annotated datasets \cite{deng2009imagenet}, visual object recognition has made remarkable achievements. However, to obtain a good multi-class classifier, thousands of high-quality labeled images are needed to train the model, the learned classifier can only identify the categories that participate in the training. In real-world applications, there are tens of thousands of categories, most of which are long tail. It is laboriously to collect and label large amounts of data for each category. Moreover, new categories are emerging endlessly, it is also difficult to collect sufficient training data for these new categories. Therefore, how to develop a robust algorithm for classification in the case of limited samples or even zero samples is a very urgent and arduous task.
\begin{figure}[t]
\centering
   \includegraphics[width=\linewidth]{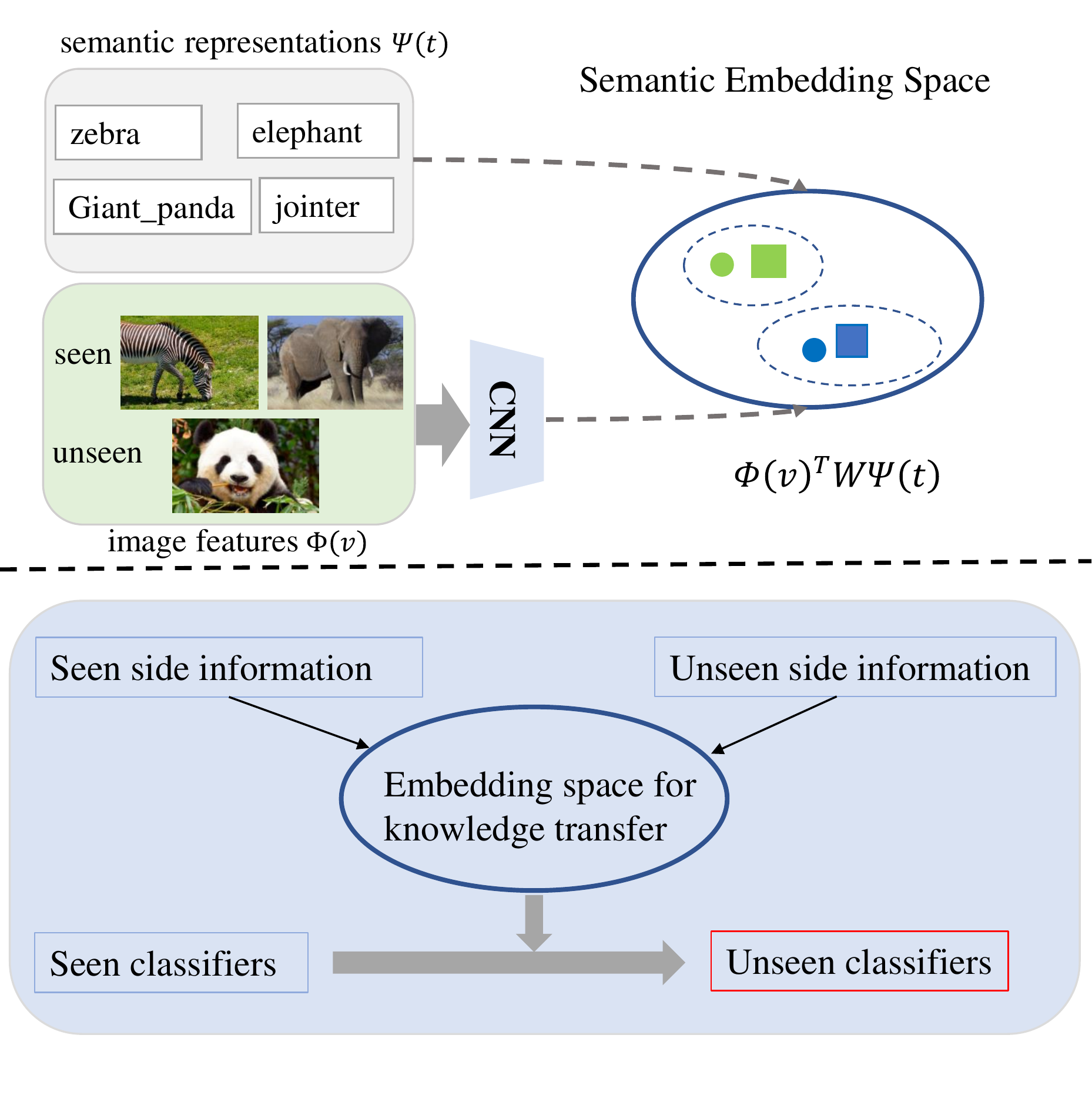}
   \caption{Illustration of the semantic embedding-based (top) and the classifiers-based (bottom) methods that are typically used for zero-shot learning.}
\label{fig1}
\end{figure}

\par To address such issues, Zero-Shot Learning (ZSL) \cite{yu2018zero, xu2019ternary, xu2019complementary, xian2018zero, yu2018transductive, hascoet2019zero, zhang2019tgg, liu2020hyperbolic, rahman2018unified, kampffmeyer2019rethinking, xu2019ternary, zhang2018zero, xu2017matrix, shen2019scalable, li2019alleviating, xu2017attribute,yang2016zero, li2019leveraging} has recently attracted great attention, which aims to recognize new categories without any exemplars. The core idea of ZSL is to emulate the intelligence of humans in transfering knowledge obtained from seen categories to describe the unseen categories. Existing ZSL methods can be divided into two paradigms: (1) embedding-based. Embedding-based \cite{li2015zero, socher2013zero, lampert2014attribute} methods learn a projection from visual space to semantic (e.g., attribute or word embedding) space based on the dataset of seen categories. In such a setting, an effective common embedding space is constructed to associate seen categories and unseen categories for knowledge transfer. (2) classifier-based. Classifier-based methods \cite{wei2019residual, changpinyo2016synthesized, wang2018zero} learn a classifier for new categories. The core idea behind these methods is to use semantic information (e.g., word embedding) to model correlations among seen and unseen categories in semantic space, and learn a classifier for unseen categories. As illustrated in Figure. \ref{fig1}. 
%As illustrate in Figure 1 (a), 

Embedding-based methods have a general hypothesis that a projection function existing in semantic space measures the compatibility between the visual features and the side information for both seen and unseen categories \cite{li2018discriminative}. During training, a mapping matrix is learned from the labeled data. At test phase, the learned mapping matrix is shared between the seen and unseen categories, so unlabeled data can be accurately identified by resorting to the nearest neighbor search in the common semantic space. In the last decade, many regularization terms and various optimization objectives have been extensively studied to obtain better mapping matrix. 
\par The alternative and less-explored paradigm is to generate a classifier for novel categories based on the explicit or implicit correlations among seen and unseen categories. In this paradigm, the key issue is to model the correlations among seen and unseen categories. Li et al. \cite{li2017zero} learned simple linear semantic correlations among seen and unseen category prototypes, and generated a classifier for unseen categories based on the learned semantic correlations. Misra et al. \cite{misra2017red} generated classifiers for complex visual concepts by composing simple concepts primitives classifiers. However, the correlations among different categories can be more complex than a linear combination. Moreover, the above methods only explored the correlations among categories in an implicit way.

\par Recently, some studies have used structured knowledge information (knowledge graph) to model explicit correlations among categories, and achieved favorable performance in various computer vision tasks such as zero-shot action recognition \cite{gao2019know, TIP2019Ji}, multi-label image classification \cite{marino2017more, chen2019multi}, cross-modal retrieval~\cite{WangYXHS17, XuSYSL17, xu2019ternary}, object detection \cite{fang2017object}, etc. The great advantage of knowledge graph is that it allows statistical strength to be shared between related categories. Due to the particular structure of the knowledge graph, to effectively utilize the knowledge information in the knowledge graph, Graph Convolutional Networks (GCN) \cite{kipfsemi, zhou2018graph} is used to propagate information among different categories. Wang et al. \cite{wang2018zero} designed a 6-layer deep GCN for zero-shot recognition.

However, deep GCN suffers from over-smoothing problem.
In recent work, Li et al. \cite{li2018deeper} demonstrated that GCN performs a form of Laplacian smoothing. As the number of GCN layers increases, the node features may be over-smoothed so that node features from different categories may become indistinguishable \cite{chen2019multi}. To alleviate the problem of over-smoothing caused by deep GCN, we design a novel Residual Graph Convolutional Network (ResGCN) for ZSL by introducing residual connections into the hidden layer of GCN.  Veit et al. \cite{veit2016residual} proved that the residual networks \cite{he2016deep} can be viewed as an ensemble of shallow networks. Residual Graph Convolutional Network has the same merit with residual networks and can effectively alleviate over-smoothing.

Furthermore, existing knowledge graph-based methods directly extract fixed relationships from WordNet \cite{miller1995wordnet}, which limits the scalability of knowledge graphs. Since WordNet is built by experts, it is difficult to absorb new categories. To address this problem, we propose a new method to build a Semantic Enhanced Knowledge Graph (SE-KG) that contains both expert knowledge and semantic information.
In practice, the word embeddings of related categories are close to each other in the semantic space. Therefore, if the word embeddings are close to each other in the semantic space, their corresponding categories should have edges connected in the knowledge graph. Concretely, we first build a basic knowledge graph from WordNet, and then connect the categories whose word embeddings are close to each other in the semantic space. For a new category that is not in WordNet, it can be absorbed by connecting it with the semantically closer category in the knowledge graph.

Most existing zero-shot methods are well designed and can achieve competitive results on small datasets. However, their performance degrades rapidly on the large-scale datasets, such as ImageNet-21K \cite{deng2009imagenet}. One possible reason could be that these methods learned an ambiguous inter-class relationship to transfer knowledge. As the number of unseen categories increases, the drawback of these methods become more serious. Furthermore, one major shortcoming of small datasets is that the number of unseen categories is dwarfed by the number of seen categories. To solve such issues, we provide a new method that focuses on the large-scale zero-shot image classification. We evaluate our method mainly on the large-scale ImageNet-21K dataset. For generalization, we also report experimental results on the AWA2 \cite{xian2018zero} dataset.

\par The main contributions of this paper are summarized as follows:
\begin{itemize}
\item We design a novel Residual Graph Convolutional Network (ResGCN) for ZSL, which can effectively alleviate the problem of over-smoothing in GCN based models.
\item We present a new method to build a semantic enhanced knowledge graph that contains both expert knowledge and semantic information. Our method can further enhance the correlations among categories and easily absorb new categories into the knowledge graph.
\item We conduct extensive experiments and evaluate our method on two widely used benchmarks, AWA2 and large-scale ImageNet-21K datasets. Experimental results show the effectiveness of our method.
\item We show that our method can be easily applied to various feature extractor architectures and has better generalization and robustness.
\end{itemize}

This paper provides a more comprehensive development of the previous work \cite{wei2019residual}, with more analysis, insights and evaluation. Compared to the previous work \cite{wei2019residual}, (1) we have strengthened the introduction of our work. In particular, we have elaborated more details about the motivation of researching on zero-shot learning and the drawback of existing work in this field. (2) in Section \ref{33}, we provide two new variants of residual convolutional networks to further improve performance. In the previous work \cite{wei2019residual}, our residual block only skipping two layers. Noted that shallow GCN may limit representation abilities. To handle this problem, we provide two new variants of residual GCN with more skipping layers. (3) the experiment part is re-designed. More comparison methods are added to comprehensively illustrate the superiority of our model. In Section \ref{4}, we provide in-depth analysis of experimental results. Specifically, (a) in Section \ref{awa}, we evaluate our method on more datasets and achieve competitive results; (b) in Section \ref{qua}, we provide Qualitative Results; (c) in Section \ref{ablation}, we provide Ablation Study with more experimental setting. Furthermore, we apply our method to various architectures. Experimental results show that the proposed method has better generalization and robustness.

\par The rest of this paper is organized as follows. Section \ref{2} provides a review of related work. In Section \ref{3}, we elaborate on the proposed residual graph convolution network and semantic enhanced knowledge graph. In Section \ref{4}, we test our method on the two benchmarks. The conclusion is presented in Section \ref{con}.

\section{Related Work}
\label{2}
In this section, we review related work on the study of zero-shot learning. Existing methods are divided into two categories: embedding-based and classifier-based.

{\bf Embedding-based methods.} Most previous work introduced linear or nonlinear modeling methods to learn a mapping matrix for zero-shot learning. Generality, these methods have two stages: firstly, learn a mapping matrix between visual features and side information (e.g., attribute, word embedding) on the training dataset; secondly, predict the test samples by performing the nearest neighbor search in the common embedding space. Attributes which defined as various properties of categories are widely used in early ZSL work. Lampert et al. \cite{lampert2014attribute} proposed a method named Direct Attributes Prediction (DAP), which infers the attributes of an image for the first time. Li et al. \cite{li2018discriminative} proposed a method to learn discriminative semantic representations in an augmented space for attributes. However, the manually-defined attributes are subjective, and attributes annotation are difficult and laborious. 

\par As an alternative, word embeddings have been widely studied for ZSL. Word embeddings are learned from a large linguistic corpora \cite{socher2013zero}, each category can be uniquely represented by a word embedding vector in the word embedding space. Socher et al. \cite{socher2013zero} proposed a non-linear model to project visual features into the word embedding space. Frome et al. \cite{frome2013devise} designed a linear transformation layer to measure the compatibility between the visual features and word embeddings. In order to explore the complementary of different semantic information, Ji et al. \cite{ji2017zero} proposed a model named Multi-Battery Factor Analysis (MBFA), which constructed a unified semantic space for both visual features and multiple types of semantic information. Instead of learning a projection function between visual features and semantic information, Changpingyo et al. \cite{changpinyo2017predicting} proposed a kernel-based method to learn an exemplar for each category through semantic representation. Norouzi et al. \cite{norouzi2014zero} directly generated a semantic vector for each unseen category via a convex combination of seen categories word embeddings. Above methods all adopt the semantic space as a bridge to measure the compatibility between the visual features and the semantic information, however, the generalization capabilities of semantic models and mapping models are limited, resulting in poor performance of these methods.

{\bf Classifiers-based methods.} Recently, there are many methods to learn a classifier for each new category. The core idea behind these methods is to mine correlations among categories and generate classifiers for new categories based on this correlation. Li et al. \cite{li2017zero} proposed a model to learn the correlations among seen and unseen prototypes, thus the unseen classifiers can be generated by combinations of seen classifiers using the same correlations. Changpinyo et al. \cite{changpinyo2016synthesized} proposed a method to synthesize classifiers for unseen categories via aligning the semantic space and visual space. Nevertheless, the above approaches only mine the implicit correlations among categories. Another popular way to transfer knowledge is using knowledge graph to model the explicit correlations among categories. Marino et al. \cite{marino2017more} proposed a graph search neural network and used knowledge graph for object recognition. Lee et al. \cite{lee2018multi} introduced a knowledge graph to model the correlations among multiple labels. One strong shortcoming of these methods is they did not incorporate unseen labels during training \cite{gao2019know}. DGP \cite{kampffmeyer2019rethinking} exploits the hierarchical structure of the knowledge graph by connecting the ancestors and descendants. Wang et al. \cite{wang2018zero} introduced a knowledge graph to model the explicit correlations among categories, and used a deep graph convolutional network to propagate message among different categories. However, this method is deficient since the knowledge graph they built only considers expert knowledge and ignores the labels semantic correlation. Moreover, deep graph convolutional network suffers from over-smoothing problem.

To address such issues, we introduce a new semantic enhanced knowledge graph (SE-KG) that contains both expert knowledge and labels semantic correlation. In addition, we design a novel residual graph convolutional network to propagate messages among different categories in the new SE-KG. The proposed residual graph convolutional network can effectively alleviate over-smoothing problem.

%-------------------------------------------------------------------------
\section{Proposed Method}
\label{3}
\subsection{Problem Statement}

Assume that there is a labeled source dataset $S={\{(\bs{x_i},y_i)\}}_{i=1}^{ms}$, where $\bs{x_i}$ denoted as the visual feature vector, and each $\bs{x_i}$ is associated with a label $y_i$. There are $N_s$ distinct categories available for source dataset, ${y_i}\in Y_s=\{1, 2, 3,...,N_s\}$. In addition, there is a target dataset $T={\{({\bs{x_j}}\prime,{y_j}\prime)\}}_{i=1}^{mt}$, where $\bs{x_j}\prime $ denoted as visual feature vector, while ${y_j}\prime \in Y_t=\{N_s+1,\cdots,N_s+N_t\}$. Target dataset contains $N_t$ distinct categories, which are disjoint with source target, $Y_s \cap Y_t= \emptyset$. In general, the categories within source dataset are also referred to as the seen categories, the categories within target dataset are referred to as the unseen categories. Zero-shot learning aims to learn a classifier $(F: \bs{x_i}\to y_i)$ on the $S$, which performs well on the target dataset $T$ $(F:{\bs{x_j}}\prime \to {y_j}\prime)$. Zero-shot learning is a non-trivial task, but it is infeasible without any side information. To mitigate zero-shot learning, in this paper, we formulated zero-shot learning as a classifier weights regression problem. Concretely, we associate each label with a word embedding, and learn a projection from word embedding to visual classifier for both seen and unseen categories.

\begin{figure*}[!htb]
\centering
\includegraphics[width=\linewidth]{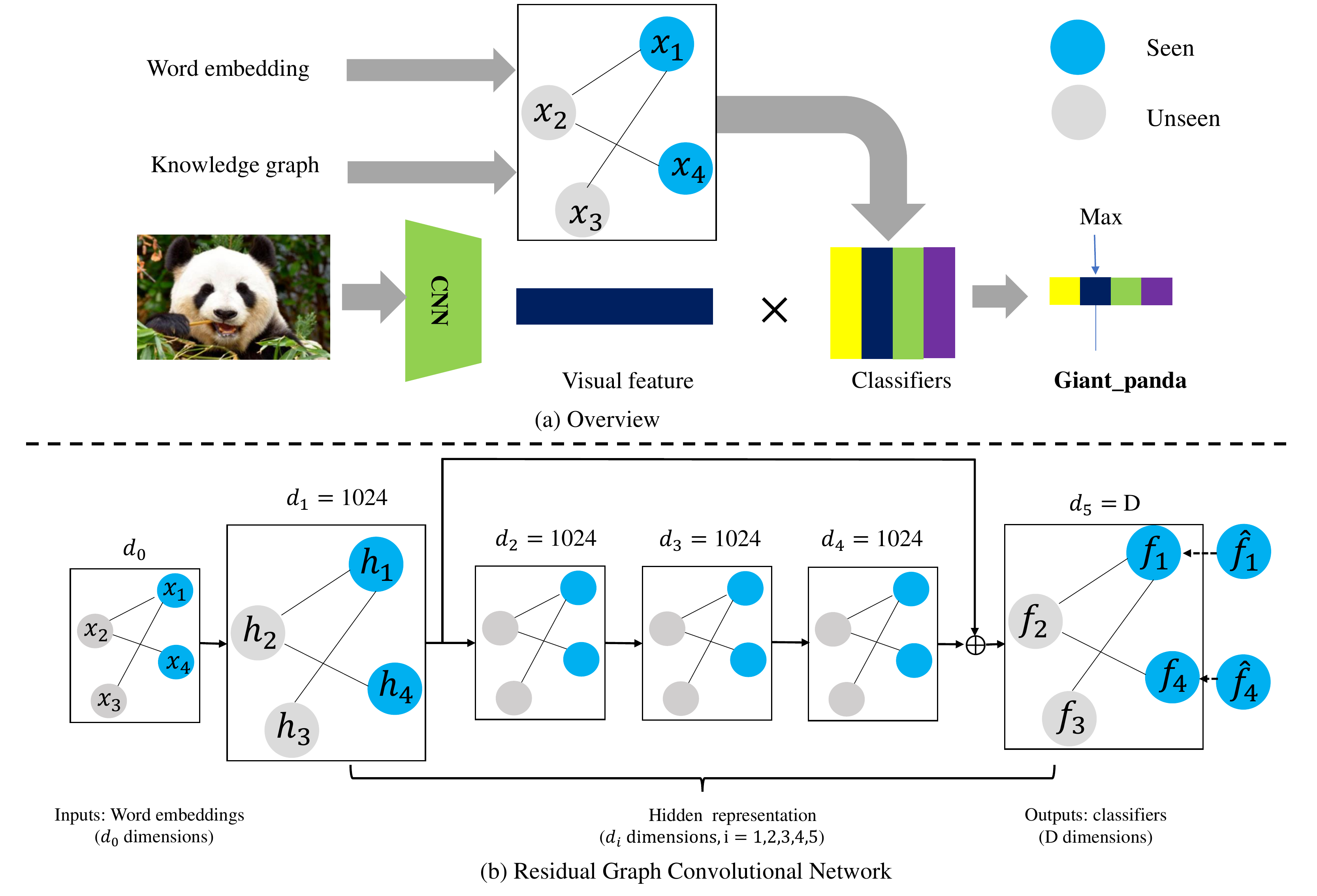}
\caption{Example of a Residual Graph Convolutional Network for zero-shot learning. The input is the matrix of word embeddings and outputs are classifiers for each category. The ground-truth $ \hat{\bs{f}_1} $ is extracted from the pre-trained CNN. During testing, with the classifiers predicted by our model, the inner product of the classifiers and the visual feature can be obtained, the classification is performed by ranking the value of the inner product.}
\label{fig:short}
\end{figure*}

\subsection{Graph Convolutional Network}
Graph Convolutional Network (GCN) \cite{kipfsemi} aims to semi-supervised learning on graph-structured data. The key idea is to propagate message between different nodes via explicit graph-based correlations among categories. In semi-supervised setting, given an undirected graph that models the explicit correlations among different entities, where each node represents an entity and related entities are connected by edges. GCN takes nodes feature matrix $\mathcal{H}^l\in \mathbb{R}^{N\times d}$ and adjacency matrix $\mathcal{A}\in \mathbb{R}^{N\times N}$ (binary or weighted) as inputs, and updates the nodes feature as $\mathcal{H}^{l+1}\in \mathbb{R}^{N\times m}$, here $\mathcal{H}^l \in \mathbb{R}^{N\times d}$ is the hidden representation of the $l$-th layer, $N$ represents the number of nodes, and $d$ denotes the dimension of node features. Formally, layer-wise propagation rule in GCN can be defined as follows:
\begin{equation}
\mathcal{H}^{l+1}=\boldsymbol{\sigma}(\hat{\mathcal{A}} \mathcal{H}^l \mathcal{W}^l ),
\label{eq1}
\end{equation}
where $\hat{\mathcal{A}}\in \mathbb{R}^{N\times N}$ is the normalized version of adjacency matrix $\mathcal{A}$. $\boldsymbol{\sigma}(\cdot)$ denotes a nonlinear activation function. $\mathcal{W}^l \in \mathbb{R}^{d \times m}$ are the parameters of the $l$-th layer.

\subsection{Residual Graph Convolutional Network for ZSL}
\label{33}
A deep GCN can be obtained by stacking multiple graph convolutional layers of the form of Eq. \ref{eq1}. In recent work, Li et al. \cite{li2018deeper} demonstrated that GCN performs a form of Laplacian smoothing.  As the number of GCN layers increases, node features from different categories may become indistinguishable \cite{chen2019multi} due to over-smoothing. To address this issue, we explore introducing residual connections into GCN. Our work is inspired by residual networks \cite{he2016deep}. More recently, Veit et al. \cite{veit2016residual} proved that the residual networks \cite{he2016deep} can be viewed as an ensemble of shallow networks, deep convolutional networks can benefit by introducing residual connections. Residual Graph Convolutional Network has the same merit with residual networks \cite{he2016deep} and can effectively alleviate over-smoothing.

\begin{table*}
%\centering
%\scriptsize
\caption{Detailed architectures for the task of ZSL. ``Identity'' means that the identity shortcut is used. ``Projection'' means using the projection shortcut to match the dimension. D represents the dimension of the ground-truth classifier.}
\label{arc}
\begin{center}
\setlength{\tabcolsep}{5mm}{
\begin{tabular}{l|c|c|c|c|c}
\hline
Layer name&Model-1&Model-2&Model-3&Model-4&Model-5 \\
\hline
First layer& 1024 &2048 &2048 &1024&1024\\
\cline{1-6}
\multirow{6}{*}{\shortstack{Residual\\Connections} }&  \multirow{3}{*}{ \shortstack{(Identity)\\1024} }  &\multirow{3}{*}{ \shortstack{(Projection)\\2048\\1024} } &\multirow{3}{*}{ \shortstack{(Identity)\\2048\\2048} }& \multirow{6}{*}{ \shortstack{(Identity) \\1024\\1024\\1024} } &\multirow{6}{*}{ \shortstack{(Identity)\\1024\\2048\\1024} }\\
%\cline{1-1}
&  & & & &\\
&&&&&\\
\cline{2-4}
& \multirow{3}{*}{ \shortstack{(Projection) \\2048}}&\multirow{3}{*}{ \shortstack{(Identity) \\1024\\1024}} &\multirow{3}{*}{ \shortstack{(Projection) \\2048\\1024}}&&\\
&  & & & &\\
&&&&&\\
\hline
Last layer&  D &D&D &D&D\\
\hline
\end{tabular}}
\end{center}
\end{table*}

Similar to the residual networks \cite{he2016deep}, we introduce residual connections to every few stacked GCN layers. There are two forms of residual connections: identity shortcuts and projection shortcuts. For clarity of notation, we use function $F(\cdot)$ to encode a few stacked GCN layers, $F(\cdot)$ in this paper has one, two or three GCN layers. Identity shortcuts can be formulated as follows:
\begin{equation}
\mathcal{X}_{out}=F(\mathcal{X}_{in},\hat{\mathcal{A}},\{\mathcal{W}_i\})+\mathcal{X}_{in},
\label{eq5}
\end{equation}
where $\hat{\mathcal{A}}$ is the normalized version of adjacency matrix $\mathcal{A}$, ${\mathcal{W}_i}$ are the parameters need to be learned, $\mathcal{X}_{in}$ and $\mathcal{X}_{out}$ are the input and output nodes feature matrix of the GCN layers considered. Note that $\mathcal{X}_{in}$ and $F(\mathcal{X}_{in},\hat{\mathcal{A}},\{\mathcal{W}_i\})$ have the same dimension. The operation\ $F(\mathcal{X}_{in},\hat{\mathcal{A}},\{\mathcal{W}_i\})+\mathcal{X}_{in}$ is performed by element-wise addition.
\par Projection shortcuts are used for case where the dimensions of $\mathcal{X}_{in}$ and $\mathcal{X}_{out}$ are not same, in which a linear projection is performed to match the dimensions. Projection shortcuts can be defined as:
\begin{equation}
\mathcal{X}_{out}=F(\mathcal{\mathcal{X}}_{in},\hat{\mathcal{A}},\{\mathcal{W}_i\})+(\hat{\mathcal{A}}\mathcal{X}_{in}\mathcal{W}_j),
\end{equation}
where $\mathcal{W}_j$ is the parameter need to be learned. Note that the dimensions of ${(\hat{\mathcal{A}}\mathcal{X}}_{in}\mathcal{W}_j)$ and $F(\mathcal{X}_{in},\hat{\mathcal{A}},\{\mathcal{W}_i\})$ are same.
\par Based on the above two forms of residual connections, we can design a residual graph convolutional network. Concretely, the identity shortcuts are used when the input and output have the same dimension. If not the case, the projection shortcuts are performed to match the dimensions. We have designed various residual graph convolutional networks for ZSL. One of our architectures is illustrated in Figure. \ref{fig:short}, various detailed architectures are presented in Table \ref{arc}. Model 1 is designed to verify the effectiveness of single-skip. Models 2 and 3 are designed to verify whether hidden layer dimensions have an impact on model performance. However, shallow GCN in the residual block may limit representation abilities. To handle this problem, we provide two new variant networks, models 4 and 5, with more skipping GCN layers. Through comparative experiments, we can obtain the best architecture for ZSL. Note that model 4 is selected for subsequent experiments. More experimental results are presented in Section \ref{ablation}.

\par  In this paper, we formulated zero-shot learning as a classifier weights regression problem. Concretely, we associate each label with a word embedding, and learn a projection from word embedding to visual classifier for both seen and unseen categories. For clarity of notation, given a dataset with $N$ categories, $S_D={\{(\bs{p}_i,\bs{f}_i)\}}_{i=1}^N$, where $\bs{p}_i$ represents the prototype of category $i$, and $\bs{f}_i$ denoted as the classifier weights. Note that each $\bs{p}_i$ is associated with a classifier $\bs{f}_i$, and $\bs{f}_i$ has the same dimension with visual features. In this paper, word embeddings are adopted as the prototypes of category. Specifically, $\bs{p}_i$ is known for all categories, and $\bs{f}_i$ is known for a subset of category. Let $S_t={\{(\bs{p}_j,\hat{\bs{f}_j})\}}_{j=1}^M \in S_D$ be the set of training data. The goal of ZSL is to predict $\bs{f}_i$ for the remaining $N-M$ categories. During training, the input to our model is a matrix $\mathcal{P}={\{\bs{p}_i\}}_{i=1}^N $ of all category word embeddings, and the output is a set of visual classifiers ${F=\{\bs{f}_i\}}_{i=1}^N$. Formally, our model can be defined as follows:
\begin{equation}
F=\sigma (\cdots \sigma (\hat{\mathcal{A}} \mathcal{P} \mathcal{W}^0) \cdots ).
\end{equation}
During training, we adopt the mean-square error loss to optimize the model.
\begin{equation}
Loss=\frac{1}{M}\sum_{j\in S_{t}} \mathit{MSE}(\bs{f}_j,\hat{\bs{f}_j}),
\label{eq4}
\end{equation}
here, $\bs{f}_j$ is the predicted classifier and $\hat{\bs{f}_j}$ is the ground-truth classifier which obtained from the top layer of a pre-trained ConvNet, such as ResNet-50. At test phase, we use the predicted classifiers to perform classification on the unseen category visual features. The predicted scores can be obtained by the Eq. \ref{eq5}:
\begin{equation}
y_i=\mathop{MAX}\limits_{\bs{f}_j \in F}\{{\bs{f}_j}^T\cdot \bs{x}_i\},
\label{eq5}
\end{equation}
where $\bs{x}_i$ is the visual features extracted from a pre-trained ConvNet, and $y_i$ is the predicted label. Therefore, unseen categories can be accurately recognized without any exemplars in the training stage.

\begin{algorithm}[!htb]
\caption{The construction process of the SE-KG.}
\label{alg1}
\SetAlgoLined
\SetKwInOut{Input}{Input}
\SetKwInOut{Output}{Output}

\Input{WordNet knowledge graph, GloVe, $\alpha$, $k$}

1: Extract the category name from WordNet\;
2: Generate the word embedding for each category via GloVe\;
3: Calculate the distance matrix of word embedding\;
4: Select $k$ categories with the smallest distance from the current category for each category\;
5: Delete neighbors with distance greater than $\alpha$\;
6: Construct a $k$-NNG\;
7: Calculate the adjacency matrix $\mathcal{B}$ of $k$-NNG\;
8: Calculate the adjacency matrix $\mathcal{A}$ of WordNet knowledge graph\;
	9: Generate the adjacency matrix $\mathcal{C}$ of SE-KG, $\mathcal{C}=\mathcal{A}+\mathcal{B}$\;
% by adding the adjacency matrix $A$ of the $k$-NNG and the adjacency matrix of WordNet knowledge graph,\;

\Output{Adjacency matrix $\mathcal{C}$ of SE-KG}

\end{algorithm}
\subsection{Semantic Enhanced  Knowledge Graph for ZSL}
%The core idea of Zero-Shot Learning is to transfer knowledge obtained from seen categories to describe unseen categories based on the correlations among categories. %residual graph convolutional network propagates information between different categories based on the knowledge graph. 
In practice, the word embeddings of related categories are close to each other in the semantic space. Therefore, if the word embeddings are close to each other in the semantic space, their corresponding categories should have edges connected in the knowledge graph. In fact, both expert knowledge and labels semantic correlation can contribute to the ZSL task either in a directly or indirectly way. To effectively use the correlations among categories, we introduce a new Semantic Enhanced Knowledge Graph (SE-KG) that contains both expert knowledge and labels semantic information. 
%To the best of our knowledge, our work associates the expert domain knowledge and semantic relations for the first time in ZSL, which also provides an inspiration for building a better knowledge graph.Semantic Enhanced Knowledge Graph (SE-KG)

\par In this section, we will elaborate on the proposed SE-KG. Firstly, we obtain a basic knowledge graph from WordNet, which contains expert knowledge. Let $\mathcal{A}$ be the adjacency matrix of the basic knowledge graph with self-connections. Secondly, generate the labels word embedding via a pre-trained language model (e.g., GloVe \cite{pennington2014glove}), and construct a $k$-nearest neighbor graph ($k$-NNG) based on the distance of word embeddings. The $k$-nearest neighbor graph ($k$-NNG) is a graph in which two vertices $p$ and $q$ are connected by an edge, if the distance between $p$ and $q$ is among the $k$-th smallest distances from $p$ to other objects. Let $\mathcal{B}$ be the adjacency matrix of the $k$-NNG. Although other distance metrics are feasible, we use Euclidean distance in this paper. Finally, the adjacency matrix of the SE-KG can be obtained by adding the adjacency matrix of the basic knowledge graph and the adjacency matrix of $k$-NNG. Let $\mathcal{C}$ be the adjacency matrix of SE-KG, $\mathcal{C}=\mathcal{A}+\mathcal{B}$.%Mark the new knowledge graph as $G_d$,
\par In practice, there are some categories weakly related to others, and their corresponding word embeddings are far away from the word embeddings of all other categories in the semantic space. To prevent these categories from being connected to their neighbor categories in the $k$-NNG, we set a threshold $\alpha$ for the distance. Concretely, for the nearest $k$ neighbor categories of a category, only the category whose distance is less than the threshold $\alpha$ will be connected to it in the $k$-NNG. Algorithm \ref{alg1} gives the detailed algorithm for the SE-KG.

\section{Experiments}
\label{4}
%To evaluate the effectiveness of the proposed method for GZSL,  
\subsection{Implementation Details}

\noindent\textbf{Datasets and Splits.} In this section, we introduce the datasets and data splits used in this paper. The detailed data splits are summarized in Table \ref{data}.

\begin{table}
\centering
%\scriptsize
\caption{Datasets and splits.}
\label{data}
\begin{center}
\setlength{\tabcolsep}{4mm}{
\begin{tabular}{l|c|c|c}
\hline
Datasets & Subset&Seen &Unseen\\
\hline
\multirow{3}{*}{ImageNet-21K} &2-hops & \multirow{3}{*}{1,000}&1,549 \\
\cline{4-4}
&3-hops&&7,860\\
\cline{4-4}
&all &&20,842\\
\cline{4-4}
\hline
AWA2&&40&10\\
\hline
\end{tabular}}
\end{center}
\end{table}

\noindent\textbf{ImageNet-21K.} Most existing ZSL datasets have one major shortcoming is that the number of unseen categories is dwarfed by the number of seen categories. However, in practice, the number of unseen categories is much larger than the seen categories. To this end, we test our approach on the large-scale ImageNet-21K \cite{deng2009imagenet} dataset, which contains a total of 21,842 categories. In order to compare with existing state-of-the-art methods, we mirror the train/test split setting of \cite{wang2018zero, frome2013devise, norouzi2014zero}. Concretely, we train our model on the ImageNet 2012 1K dataset with 1,000 seen categories, and test on three datasets with increasing degree of difficulty. Three datasets are extracted from the large-scale ImageNet-21K dataset based on the semantic similarity to the training set labels. “2-hops” consists of roughly 1.5K categories, which within 2 tree hops of the ImageNet 2012 1K categories. “3-hops” consists of roughly 7.8K categories, which within 3 tree hops of the ImageNet 2012 1K labels. “All” includes all the ImageNet-21K labels. Noted that the labels in three datasets are disjoint with the ImageNet 2012 1K labels. 

\noindent\textbf{AWA2.} AWA2 \cite{xian2018zero} is an extension version of the original Animals with Attributes dataset and contains 37,332 images of 50 categories. We following the data split in \cite{xian2018zero} to ensure that the test categories are disjoint with the ImageNet 2012 dataset. In  \cite{xian2018zero}, 40 categories are used for training and 10 for testing. Note that the AWA2 test categories are contained in the ImageNet-21K categories, we report result on the AWA2 by using the same approach with the ImageNet-21K dataset without extra training.

\noindent\textbf{Zero-Shot Learning Settings.} There are two zero-shot test settings: Conventional setting and Generalized setting. In conventional setting, at test phase, candidate categories being only unseen categories (without training labels); In generalized setting, candidate categories include both seen and unseen labels. Note that both conventional setting and generalized setting are tested on the same datasets, the difference is in the number of candidate labels. We report experimental results in both test settings.

\begin{table*}
%\centering
%\scriptsize
\begin{center}
\caption{Hit@k performance for the different models on ImageNet zero-shot learning task. Testing on unseen categories. $\ddagger$ indicated models without fine-tuning. }
\label{table1}
\setlength{\tabcolsep}{4mm}{
\begin{tabular}{l|l|l|ccccc}
\hline
\multirow{2}{*}{Test Set} &\multirow{2}{*}{Model} & \multirow{2}{*}{Reference} & \multicolumn{5}{c}{Hit@k (\%)} \\
\cline{4-8}
& & &1 & 2& 5& 10 &20\\
\hline
\multirow{9}{*}{2-hops} & ConSE \cite{norouzi2014zero} &ICLR2014& 8.3 & 12.9 & 21.8 & 30.9 & 41.7  \\
&SYNC \cite{changpinyo2016synthesized}& CVPR2016&10.5& 17.7&28.6&40.1 & 52.0 \\
&PSR \cite{annadani2018preserving}& CVPR2018&9.4 &- &26.3&-&-\\
&EXEM \cite{changpinyo2017predicting}& ICCV2017&12.5 & 19.5 & 32.3 & 43.7 & 55.2\\
&GCNZ \cite{wang2018zero}& CVPR2018&19.8& 33.3 & 53.2& 65.4& 74.6\\
&DGP(-f)$^\ddagger$ \cite{kampffmeyer2019rethinking}& CVPR2019&23.95& 37.04 & 56.53& 69.30& 78.75\\
&Cacheux et al. \cite{cacheux2019modeling}&ICCV2019 &9.81 &- &-&-&-\\
&Zhu et al. \cite{zhu2019learning}&ICCV2019&11.0 &- &-&-&-\\
%&DGP \cite{kampffmeyer2019rethinking}&ResNet-50&23.8&36.9&56.2&69.1&78.6\\
\cline{2-8}
%& Ours& ResNet-50&24.45& 38.01& 57.20 &69.60&78.95 \\
%& GCNZ ($G_d$)& ResNet-50&\bfseries{22.43}& \bfseries{34.87}& \bfseries{54.21} &\bfseries{67.20}&\bfseries{77.15} \\
& \bfseries{ResGCN} &\bfseries{Ours}&\bfseries{26.68}& \bfseries{40.59}& \bfseries{60.25} &\bfseries{71.73}&\bfseries{80.68} \\
\hline
\multirow{9}{*}{3-hops}&ConSE \cite{norouzi2014zero}&ICLR2014& 2.6 & 4.1 & 7.3& 11.1& 16.4\\
&SYNC \cite{changpinyo2016synthesized}&CVPR2016& 2.9& 4.9& 9.2& 14.2& 20.9\\
&PSR \cite{annadani2018preserving}& CVPR2018&2.8 &- &4.8&-&-\\
&EXEM \cite{changpinyo2017predicting}&ICCV2017& 3.6& 5.9& 10.7& 16.1& 23.1\\
&GCNZ \cite{wang2018zero}& CVPR2018&4.1 & 7.5& 14.2& 20.2& 27.7\\
&DGP(-f)$^\ddagger$ \cite{kampffmeyer2019rethinking}& CVPR2019&5.67& 9.62 & 17.55& 25.60& 35.14\\
%&DGP \cite{wang2018zero}& ResNet-50&6.30& 10.70 & 19.30& 27.2& 18.38\\
&Cacheux et al. \cite{cacheux2019modeling}&ICCV2019 &2.52 &- &-&-&-\\
&Zhu et al. \cite{zhu2019learning}&ICCV2019 &2.40 &- &-&-&-\\
\cline{2-8}
%&Ours &ResNet-50& 5.70&9.97 &17.71&25.90& 34.48\\
%&GCNZ ($G_d$) &ResNet-50& \bfseries{4.58}&\bfseries{8.09} &\bfseries{15.43}& \bfseries{23.12}& \bfseries{32.23}\\
& \bfseries{ResGCN} &\bfseries{Ours}& \bfseries{6.02}&\bfseries{10.12} &\bfseries{18.35}& \bfseries{26.20}& \bfseries{35.24}\\

\hline
\multirow{9}{*}{All}& ConSE \cite{norouzi2014zero}&ICLR2014& 1.3 & 2.1 &3.8 &5.8 &8.7\\
&SYNC \cite{changpinyo2016synthesized}& CVPR2016&1.4& 2.4& 4.5& 7.1& 10.9\\
&PSR \cite{annadani2018preserving}& CVPR2018&1.0 &- &2.7&-&-\\
&EXEM \cite{changpinyo2017predicting}&ICCV2017&1.8 & 2.9& 5.3& 8.2& 12.2\\
&GCNZ \cite{wang2018zero}& CVPR2018&1.8& 3.3& 6.3& 9.1& 12.7\\
&DGP(-f)$^\ddagger$ \cite{kampffmeyer2019rethinking}& CVPR2018&2.65& 4.54 & 8.42& 12.71& 18.38\\
&Cacheux et al. \cite{cacheux2019modeling}&ICCV2019 &1.09 &- &-&-&-\\
&Zhu et al. \cite{zhu2019learning}&ICCV2019 &1.00 &- &-&-&-\\
\cline{2-8}
%& Ours &ResNet-50&2.65 &4.63 &8.30&12.30 &17.58\\
%&GCNZ ($G_d$) &ResNet-50& \bfseries{2.06}&\bfseries{3.66} &\bfseries{7.12}& \bfseries{10.48}& \bfseries{15.83}\\
& \bfseries{ResGCN} &\bfseries{Ours}& \bfseries{2.78}&\bfseries{4.74} &\bfseries{8.76}& \bfseries{12.86}& \bfseries{18.46}\\

\hline
\end{tabular}}
\end{center}
\end{table*}

\noindent\textbf{Model Details.} We adopt the GloVe \cite{pennington2014glove} model trained on the Wikipedia dataset to obtain the word embeddings of labels. The dimension of word embeddings is set at 300. If category names contain multiple words, we obtain the category word embedding as average of embeddings for all words. For activation function, we adopt LeakyReLU \cite{Bing2015Empirical} with the negative slope of 0.2. Following \cite{wang2018zero}, we perform L2-Normalization on the predicted classifiers and ground-truth classifiers. For image representations, we adopt ResNeXt-101 pre-trained on the ImageNet 2012 1K dataset to generate the visual representations. For each image, the 2048-dimensional activations of the penultimate layer are taken as visual features. We also adopt the Dropout in each graph convolutional layer, with a dropout rate of 0.5. Different from \cite{wang2018zero}, which performed symmetric normalization (Sym, $\hat{\mathcal{A}}=\mathcal{D}^{-\frac{1}{2}}\mathcal{A}\mathcal{D}^{-\frac{1}{2}}$) on adjacency matrix, we adopt a random walk normalization (Non-sym, $\hat{\mathcal{A}}=\mathcal{D}^{-1}\mathcal{A}$) on adjacency matrix. We train all architecture from scratch using Adam \cite{Kingma2014Adam} with the initial learning rate is 0.001, and a weight decay 0.0005. The model is trained for 300 epochs in total. We implement our architecture by PyTorch \cite{paszke2017automatic}.
\begin{table}
\caption{Top-1 results for unseen categories on AWA2. Results for ConSE, Devise and SYNC obtained from \cite{xian2018zero}.}
\label{awa2}
\begin{center}
\setlength{\tabcolsep}{3mm}{
\begin{tabular}{l|l|c}
\hline
Model&Reference&Hit@1(\%)\\
\hline
ConSE & ICLR2014&44.5\\
Devise&NeurIPS2013&59.7\\
SYNC& CVPR2016&46.6\\
Xian et al. \cite{xian2018feature}&CVPR2018&68.2\\
Cacheux et al. \cite{cacheux2019modeling}&ICCV2019&67.9\\
GFZSL \cite{verma2017simple}&ECML2017&63.8\\
SE-GZSL \cite{kumar2018generalized}&CVPR2018 &69.2\\
%Gaussian-Ort \cite{zhang2018zero} & 70.5\\

GCNZ  & CVPR2018&70.7\\
\hline
%Ours (ResNet-50) && 74.40\\
\bfseries{ResGCN} &\bfseries{Ours}& \bf{77.4}\\
\hline
\end{tabular}}
\end{center}
\end{table}

\subsection{Evaluation Results with Conventional Setting}
\label{awa}
\noindent\textbf{Results on ImageNet-21K.} We first contrast our approach to several existing state-of-the-art methods without considering the training labels. DeViSE~\cite{frome2013devise} learned a linearly mapping matrix from visual space to word vector space. The model is initialized by two pre-trained neural network models, and trained by a hinge ranking loss. At test phase, an image is classified by comparing its similarity with the unseen categories in the semantic embedding space. Instead of learning a mapping matrix from visual space to semantic space, ConSE \cite{norouzi2014zero} directly generated a semantic embedding vector for an image in the semantic space through the convex combination of the top-$k$ likely train labels semantic embedding vectors. A test image is classified by comparing its embedding vector with unseen categories in the semantic embedding space.

\begin{table*}
	%\centering
	%\scriptsize
\begin{center}
\caption{Hit@k performance for the different models on ImageNet zero-shot learning task. Testing on both seen and unseen categories.  $\ddagger$ indicated models without fine-tuning. }
\label{table2}
\setlength{\tabcolsep}{4mm}{
\begin{tabular}{l|l|l|ccccc}
\hline
\multirow{2}{*}{Test Set} &\multirow{2}{*}{Model} & \multirow{2}{*}{Reference} & \multicolumn{5}{c}{Hit@k (\%)} \\
\cline{4-8}
& & &1 & 2& 5& 10 &20\\
\hline
\multirow{8}{*}{2-hops+1K} & DeViSE \cite{frome2013devise} &NeurIPS2013& 0.8 & 2.7 & 7.9 & 14.2 & 22.7  \\
&ConSE \cite{norouzi2014zero}& ICLR2014&0.3& 6.2&17.0&24.9 & 33.5 \\
&PSR \cite{annadani2018preserving}& CVPR2018&1.2 &- &11.2&-&-\\
%&EXEM[23]& Inception-v1&12.5 & 19.5 & 32.3 & 43.7 & 55.2\\
& GCNZ \cite{wang2018zero}& CVPR2018&9.7& 20.4 & 42.6& 57.0& 68.2\\
%&GCNZ (SE-KG) &ResNet-50& \bfseries{11.00}&22.92 &43.62& 58.02& 70.00\\
&DGP(-f)$^\ddagger$ \cite{kampffmeyer2019rethinking}& CVPR2019&9.76& 22.61 & 44.82& 60.42& 72.24\\
&FGZSL \cite{xian2018feature}&CVPR2018 &4.40 &- &-&-&-\\
&Zhu et al. \cite{zhu2019learning}&ICCV2019 &4.50 &- &-&-&-\\
\cline{2-8}
%&Ours& ResNet-50&10.70& 24.68& 47.85 &61.99& 72.50\\
& \bfseries{ResGCN} &\bfseries{Ours}& \bfseries{10.80}&\bfseries{25.47} &\bfseries{50.41}& \bfseries{64.44}& \bfseries{75.22}\\
\hline
\multirow{8}{*}{3-hops+1K}&DeViSE \cite{frome2013devise}&NeurIPS2013& 0.5 & 1.4 & 3.4& 5.9& 9.7\\
%&SYNC&Inception-v1& 2.9& 4.9& 9.2& 14.2& 20.9\\
& ConSE \cite{norouzi2014zero}&ICLR2014& 0.2& 2.2& 5.9& 9.7& 14.3\\
&PSR \cite{annadani2018preserving}& CVPR2018&1.8 &- &4.9&-&-\\
& GCNZ \cite{wang2018zero}& CVPR2018&2.2 & 5.1& 11.9& 18.0& 25.6\\
%&GCNZ (SE-KG) &ResNet-50& \bfseries{3.03}&5.93 &13.08& 20.77& 29.95\\DGP(-f)$^\ddagger$ \cite{kampffmeyer2019rethinking}, FGZSL \cite{xian2018feature}, Zhu et al. \cite{zhu2019learning}
&DGP(-f)$^\ddagger$ \cite{kampffmeyer2019rethinking}& CVPR2019&2.81& 6.07 & 14.37& 22.72& 32.73\\
&FGZSL \cite{xian2018feature}&CVPR2018 &1.18 &- &-&-&-\\
&Zhu et al. \cite{zhu2019learning}&ICCV2019 &1.20 &- &-&-&-\\
\cline{2-8}
%&Ours &ResNet-50& 2.84&6.34 &14.53& 22.86& 32.76\\

& \bfseries{ResGCN} &\bfseries{Ours}& \bfseries{2.85}&\bfseries{6.36} &\bfseries{15.30}& \bfseries{23.50}& \bfseries{32.85}\\
\hline
\multirow{8}{*}{All+1K}& DeViSE \cite{frome2013devise}&NeurIPS2013& 0.3 & 0.8 &1.9 &3.2 &5.3\\
%&SYNC & Inception-v1&1.4& 2.4& 4.5& 7.1& 10.9\\
&ConSE \cite{norouzi2014zero}&ICLR2014&0.2 & 1.2& 3.0& 5.0& 7.5\\
&PSR \cite{annadani2018preserving} & CVPR2018&0.4 &- &1.7&-&-\\
&GCNZ \cite{wang2018zero}& CVPR2018&1.0& 2.3& 5.3& 8.1& 11.7\\
%&GCNZ (SE-KG) &ResNet-50& 1.40&2.76 &6.15& 9.89& 14.90\\
&DGP(-f)$^\ddagger$ \cite{kampffmeyer2019rethinking}& CVPR2019&1.36& 3.00 & 7.05& 11.36& 16.10\\
&FGZSL \cite{xian2018feature}&CVPR2018 &0.45 &- &-&-&-\\
&Zhu et al. \cite{zhu2019learning}&ICCV2019 &0.50 &- &-&-&-\\
\cline{2-8}
%& Ours &ResNet-50&1.40 &3.00 &7.01 &11.36 &16.36\\
& \bfseries{ResGCN} &\bfseries{Ours}& \bfseries{1.40}&\bfseries{3.10} &\bfseries{7.17}& \bfseries{11.42}& \bfseries{16.73}\\
\hline
\end{tabular}}
\end{center}
\end{table*}

\par SYNC~\cite{changpinyo2016synthesized} predicted classifiers for unseen categories by a linear combination of the classifiers of ``phantom'' categories. EXEM~\cite{changpinyo2017predicting} predicted a visual exemplar for each unseen category in the semantic space. PSR \cite{annadani2018preserving} developed an objective function to preserve the inter-class relationship in the embedding space. The above methods all adopt the semantic space as a bridge to learn the correlations among categories, and transfer knowledge based on this correlation from seen categories to unseen categories. However, these methods did not explore explicit correlations among categories, which severely limits the performance of the model. GCNZ~\cite{wang2018zero} first used knowledge graph to model the explicit correlations among categories, and predicted logistic classifier weights for new category, which is the approach most related to us. DGP \cite{kampffmeyer2019rethinking} exploits the hierarchical structure of the knowledge graph by connecting the ancestors and descendants. DGP \cite{kampffmeyer2019rethinking} uses fine-tuning technology, which is extremely time-consuming on the large-scale dataset. For a fair comparison, we retrained DGP and report its result without fine-tuning. Zhu et al. \cite{zhu2019learning} proposed a translator network that translates the class-level semantic features to visual features.  Cacheux et al. \cite{cacheux2019modeling} proposed a model that takes into account both inter-class and intra-class relations.

\par We first evaluate our model without considering the training labels on the ImageNet-21K dataset. Experimental results are summarized in Table \ref{table1}. From Table~\ref{table1}, we have the following observations: 
\begin{itemize}
\item Our method outperforms the existing state-of-the-art methods on all datasets. There are two main reasons for this phenomenon. On one hand, our proposed residual graph convolutional network well addresses the problem of over-smoothing. On the other hand, our proposed new SE-KG further enhances the correlations among categories. 
\item Specifically, on the 2-hops dataset, our final model performs better than the DGP \cite{kampffmeyer2019rethinking} 2.73\% on top-1 accuracy. On the 3-hops dataset, our final model outperforms the GCNZ \cite{wang2018zero} by 1.92\% on top-1 accuracy. The performance gap between our method and baseline shows the effectiveness of our method.
%We can notice that the gain is diminishing as the number of unseen categories in the dataset increases.
\end{itemize}

\noindent\textbf{Results on AWA2.} We show experimental results of the proposed method on AWA2 dataset and comparisons with the existing state-of-the-art methods in Table \ref{awa2}. From the result, we can find that the proposed method outperforms all state-of-the-art methods, which again demonstrates the effectiveness of the proposed method. Note that our method is different from the baselines as it doesn't use the attributes.

\subsection{Evaluation Results with Generalized ZSL Setting}
To evaluate our method in a more general and practical setting, we following the suggestions in \cite{wang2018zero}, and report the experimental results when including both seen and unseen categories labels during testing. For the baselines, we considers DeViSe~\cite{frome2013devise}, ConSE \cite{norouzi2014zero}, PSR \cite{annadani2018preserving}, GCNZ~\cite{wang2018zero}, DGP \cite{kampffmeyer2019rethinking}, FGZSL \cite{xian2018feature} and Zhu et al. \cite{zhu2019learning}. Experimental results are summarized in Table \ref{table2}. Compared to the conventional setting (without training labels), we can see that the performance of all methods has degraded. Such degradation can be alleviated by introducing a detector, which can recognize whether an image belongs to either the seen or unseen group \cite{mandal2019out}. Note that other embedding-based methods mainly use semantic space as a bridge to transfer knowledge. However, the correlations among categories are uncertain and ambiguous in the semantic space. In our case, the correlations among categories are modeled by a structured knowledge graph, and the correlations among categories are certain and explicit. Moreover, our model consistently outperforms the other approaches on all three datasets, experimental results show that our method is more suitable for transfering knowledge from seen categories to unseen categories.

\subsection{Qualitative Results}
\label{qua}
\begin{figure*}[!htb]
\centering
%\fbox{\rule{0pt}{2in} \rule{0.9\linewidth}{0pt}}
   \includegraphics[width=0.98\textwidth]{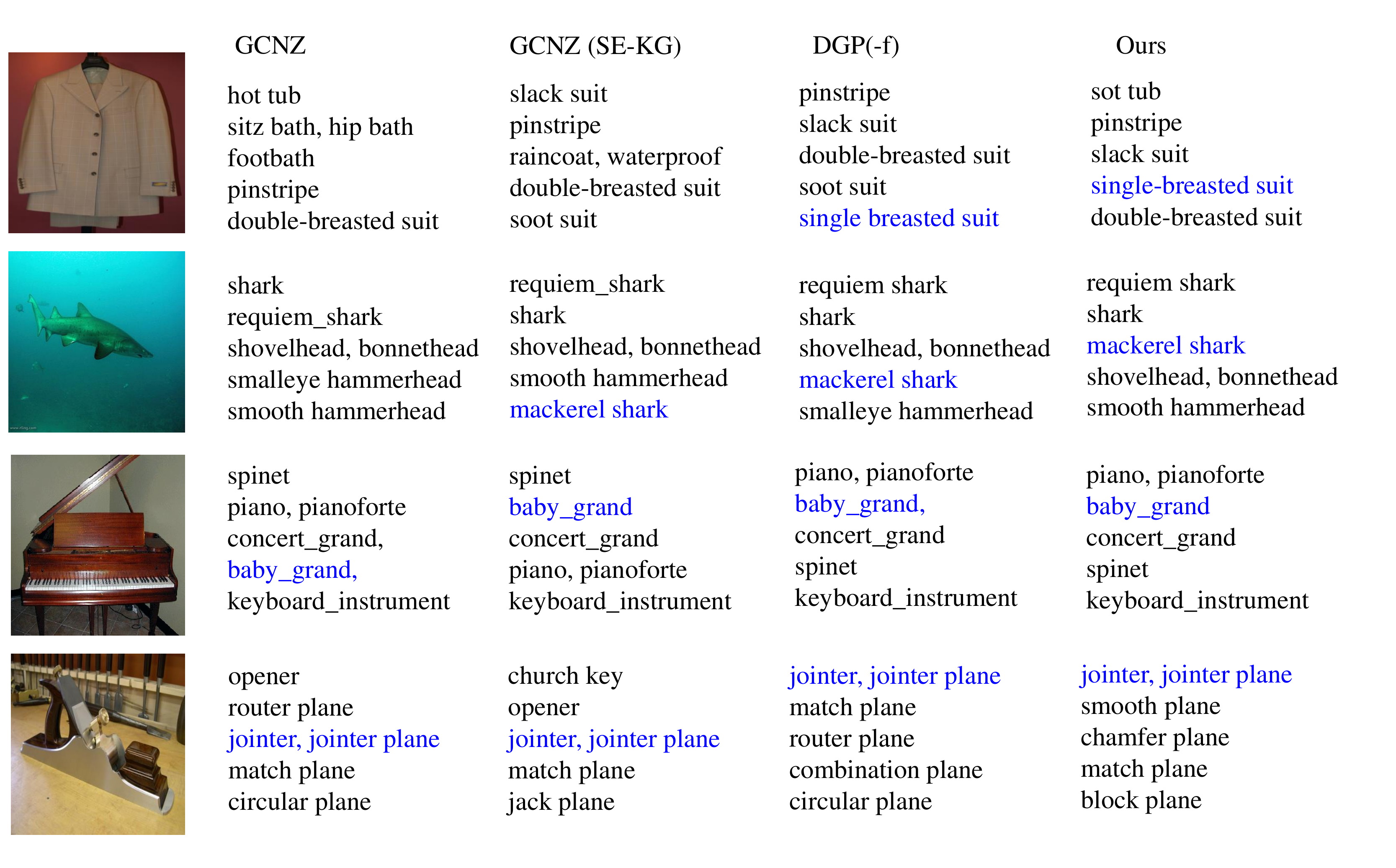}

   \caption{Top-5 prediction for 4 different images from the ImageNet-21k dataset. Predictions ordered by decreasing score, with correct labels are shown in blue.}
%\Description{Visualization of Top-5 prediction results for 4 different images.}
\label{example}
\end{figure*}

In this section, we perform visualizations Top-5 prediction results for a handful of examples from the ImageNet-21K dataset using GCNZ~\cite{wang2018zero}, GCNZ (SE-KG), DGP(-f)$^\ddagger$ \cite{kampffmeyer2019rethinking}, and ours in Figure. \ref{example}. Figure. \ref{example} qualitatively illustrates model behavior. Predictions ordered by decreasing score, with correct labels shown in blue. From these results, we find that Ours can correctly identify the samples that were misidentified by the DGP(-f)$^\ddagger$ \cite{kampffmeyer2019rethinking}, which shows the superiority of our architecture.

\subsection{Ablation Study}
\label{ablation}
\noindent\textbf{Experiments with Different Variants}. In this section, we report experimental results of different variants of residual graph convolutional network on the 2-hops dataset. For fair comparison, we report results on the WordNet knowledge graph, the pretrained ResNet50 is used to extract visual features, and word embeddings are obtained using the GloVe model. Table \ref{retarc} summarizes our experimental results and compares to the GCNZ \cite{wang2018zero}, which adopted plain deep GCN. 
% After determining the best architecture, we give the experimental results of using Resnext101 as visual feature extractor.

\begin{table}
\caption{Experimental results for distinct variants on the 2-hops dataset. For fair comparison, all results are reported with ResNet50 and WordNet knowledge graph.}
\label{retarc}
\begin{center}
\begin{tabular}{l|ccccc}
\hline
\multirow{2}{*}{Methods}& \multicolumn{5}{c}{Hit@k (\%)} \\
\cline{2-6}
%\hline
&1&2&5&10&20\\
\cline{1-6}
GCNZ \cite{wang2018zero}& 19.80& 33.30& 53.20&65.40&74.60 \\
\cline{1-6}
Model1& 20.28& 31.53& 47.62&57.42 &65.63  \\
Model2& 24.03& 37.35& 56.76&68.65 &77.76  \\
Model3& 23.91& 37.29& 56.77&68.63 &77.71  \\
Model4& \bfseries{24.21}&\bfseries{37.36} &\bfseries{56.87} &\bfseries{68.86} &\bfseries{78.26}  \\
Model5& 24.20& 37.30& 56.68&68.58 &78.06  \\
%\cline{1-6}
%Model4-ResNeXt& \bfseries{26.45}&\bfseries{40.39} &\bfseries{60.04} &\bfseries{71.78} &\bfseries{80.32}  \\
\hline
\end{tabular}
\end{center}
\end{table}

\begin{table}
\begin{center}
\caption{Hit@1 accuracies of the SE-KG with different $k$ and threshold $\alpha$. Testing on the 2-hops dataset.}
\label{para}
\setlength{\tabcolsep}{3mm}{
\begin{tabular}{c|cccc}
\hline
\diagbox{$k$}{$\alpha$}& 0.3&0.5&1.0 & 5.0 \\ 
\hline
2&26.16&{\bfseries26.68}&26.33&26.24\\
5& 26.20&26.26& 26.04&25.89\\
%\cline{2-2}
8&26.30&26.48&26.50&26.25\\
10&26.00&26.20&25.76&25.82\\

\hline
\end{tabular}}
\end{center}
\end{table}

\par The following obversions can be made from Table \ref{retarc}. First, compared to GCNZ \cite{wang2018zero}, our residual GCN has a significant performance improvement. We can see that our model 4 outperforms the GCNZ by 4.41\% on top-1 accuracy. This indicates that the over-smoothing problem is well alleviated and we manage to get performance gains by introducing residual connections. Second, we have not observed advantages for skipping single layer. For the other two connection modes, the hidden layer dimensions have a slight impact on the performance of the model. Third, by increasing the skipping layer of the residual block, the model performance is further improved. Note that adding more GCN layers in the residual block may suffer from over-smoothing problem, three GCN layers are appropriate.

\noindent\textbf{Effectiveness of SE-KG}.
There are two parameters in our SE-KG, $k$ and $\alpha$. We use grid search to determine the value of parameters. The value of $k$ is search over \{2, 5, 8, 10\} and $\alpha$ is search over \{0.3, 0.5, 1.0, 5.0\}. Experimental results are summarized in Table \ref{para}. From Table \ref{para}, we observe that our model achieves the best results at $k=2$ and $\alpha=0.5$.

\begin{figure}
\centering
%\fbox{\rule{0pt}{2in} \rule{0.9\linewidth}{0pt}}
   \includegraphics[width=\linewidth]{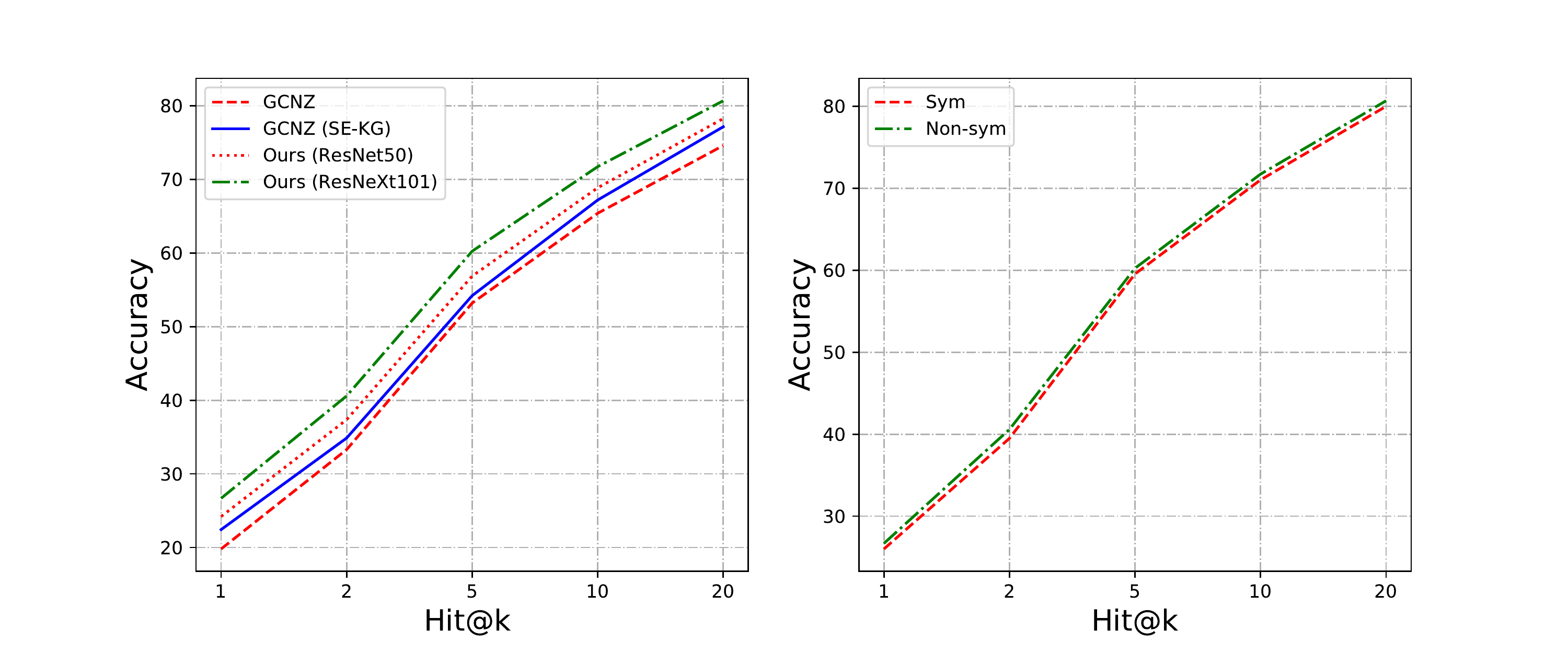}
   \caption{Left: Effectiveness of SE-KG. Right: Effect of the different normalization methods on adjacency matrix. Testing on the 2-hops dataset.}
\label{fig4}
\end{figure}

%our intuition based on the fact that if two categories are related, their word embeddings are close to each other in semantic space.if a category is not in the SE-KG, it can be easily added to the knowledge graph by connecting to its closer categories.

\par To clear evaluate the performance of SE-KG, we keep the architecture unchanged and retrain GCNZ with our new SE-KG. Experimental results are summarized in Figure. \ref{fig4}. From Figure. \ref{fig4}, we observe that retrained GCNZ (SE-KG) outperform the original model on the all tasks. This indicates that our new SE-KG can further enhance the correlations among categories, and more efficiently propagate information. Since WordNet knowledge graph is built by experts, it is difficult to absorb new categories. Our proposed SE-KG contains both expert knowledge and semantic information, which can easily add new categories to the knowledge graph by connecting new categories to its closer categories. This is a huge advantage of our method.

\noindent\textbf{Symmetric vs. Non-sym Normalization.} In this section, we report the experimental result of two normalization methods, Sym, $\hat{\mathcal{A}}=\mathcal{D}^{-\frac{1}{2}}\mathcal{A}\mathcal{D}^{-\frac{1}{2}}$ and Non-sym, $\hat{\mathcal{A}}=\mathcal{D}^{-1}\mathcal{A}$. Experimental results are summarized in Figure. \ref{fig4}. It can be seen from the results that the difference is negligible.

\noindent\textbf{Generalization and Robustness}.
To comprehensively evaluate the generalization and robustness of our method, we report results with different feature extractors. Concretely, we keep the architecture unchanged and retrained our model with different feature extractors, in this paper, we evaluate our model with the widely-used pre-trained networks including: VGG-19 \cite{simonyan2014very}, Inception-V3 \cite{szegedy2016rethinking}, DenseNet121 \cite{huang2017densely}, EfficientNet-b0 \cite{tan2019efficientnet}, EfficientNet-b7 \cite{tan2019efficientnet}, ResNet50 \cite{he2016deep}, ResNeXt101 \cite{xie2017aggregated}. During training, we extract the classifier weights from the last layer of a pre-trained CNN as the supervisor. At test time, the backbone network corresponding to the classifier is used to extract visual features. Experimental results are summarized in Table \ref{table8}. It can be seen from the results that our model shows better generalization ability and robustness on different architectures.

\begin{table}
\begin{center}
\caption{Hit@k performance for the retrained models. Testing on unseen categories.}
\label{table8}
\begin{tabular}{l|l|ccc}
\hline
\multirow{2}{*}{Test Set} &\multirow{2}{*}{Backbone Network}  & \multicolumn{3}{c}{Hit@k (\%)} \\
\cline{3-5}
 & &1 & 5& 10 \\
\hline
\multirow{7}{*}{2-hops} &VGG-19& 21.99&52.87 & 65.28\\
&ResNet50& 24.45& 57.20& 69.60\\
&Inception-v3       & 24.46& 57.31 & 68.82 \\
&DenseNet-121   & 24.68& 58.16 & 70.23 \\
&EfficientNet-b0& 24.70& 57.94 & 69.87 \\
&EfficientNet-b7& 25.96& 59.88 & 71.25 \\
& ResNeXt101 & \bfseries{26.68} &\bfseries{60.25} & \bfseries{71.73}   \\ 
\cline{1-5}
\multirow{7}{*}{3-hops}&VGG-19 & 4.80 & 15.29& 22.26\\
&ResNet50& 5.70& 17.71 & 25.90 \\
&Inception-v3       & 5.53& 17.42 & 24.91 \\
&DenseNet-121   & 5.56& 17.36 & 24.96 \\
&EfficientNet-b0& 5.59& 17.33 & 24.77 \\
&EfficientNet-b7& 5.89& 18.31 & 26.15 \\
& ResNeXt101   &  \bfseries{6.02} &\bfseries{18.35} & \bfseries{26.20} \\
%\hline
\cline{1-5}
\multirow{7}{*}{All}& VGG-19  &2.22 &7.14&10.65 \\
&ResNet50   & 2.65& 8.30 & 12.30  \\
&Inception-v3       & 2.51& 8.16 & 12.00 \\
&DenseNet-121   & 2.59& 8.20 & 12.04 \\
&EfficientNet-b0& 2.55& 8.14 & 11.91 \\
&EfficientNet-b7& 2.70& 8.68 & 12.86 \\
& ResNeXt101 &   \bfseries{2.78} &\bfseries{8.76} & \bfseries{12.86} \\
\hline
\end{tabular}
\end{center}
\end{table}

%\begin{table*}[!htb]
%	\centering
%	\scriptsize
%\begin{center}
%\caption{Hit@k performance for the retrained models on ImageNet zero-shot learning task. Testing on unseen categories.}
%\label{table8}
%\begin{tabular}{l|ccc|ccc|ccc}
%\hline
%\multirow{2}{*}{Backbone Network} &\multicolumn{3}{c}{2-hops}  & \multicolumn{3}{c}{3-hops}& \multicolumn{3}{c}{all} \\
%\cline{2-10}
%&Hit@1&Hit@5 &Hit@10&Hit@1&Hit@5 &Hit@10&Hit@1&Hit@5 &Hit@10\\
%\hline
%VGG-19& 21.99&52.87 & 65.28& 4.8 & 15.29& 22.26&2.22 &7.14&10.65\\
%ResNet50& 24.45& 57.20 &69.60& 5.70& 17.71 & 25.90 & 2.65& 8.30 & 12.30  \\
%Inception-v3       & 24.46& 57.31 & 68.82& 5.53& 17.42 & 24.91& 2.51& 8.16 & 12.00 \\
%DenseNet-121   & 24.68& 58.16 & 70.23& 5.56& 17.36 & 24.96& 2.56& 8.20 & 12.04 \\
%EfficientNet-b0& 24.70& 57.94 & 69.87& 5.59& 17.33 & 24.77& 2.55& 8.14 & 11.91 \\
%EfficientNet-b7& 25.96& 59.88 & 71.25& 5.89& 18.31 & 26.15 & 2.70& 8.68 & 12.86 \\
%ResNeXt101 & \bfseries{26.68} &\bfseries{60.25} & \bfseries{71.73} &  \bfseries{6.02} &\bfseries{18.35} & \bfseries{26.20}&   \bfseries{2.78} &\bfseries{8.76} & \bfseries{12.86} \\

%\hline
%\end{tabular}
%\end{center}
%\end{table*}

\section{Conclusion}
\label{con}

We have developed a novel Residual Graph Convolutional Network for zero-shot learning by introducing the residual connections into the hidden GCN layers. Our model can effectively alleviate the problem of over-smoothing. Moreover, we introduce a new semantic enhanced knowledge graph that contains both expert knowledge and labels semantic information. Our SE-KG can further enhance the correlations among categories and easily absorb new categories into the knowledge graph. Experimental results on two widely used benchmarks have demonstrated the effectiveness of our method. In future work, we would like to investigate the potential of more advanced knowledge graph for ZSL.

% if have a single appendix:
%\appendix[Proof of the Zonklar Equations]
% or
%\appendix  % for no appendix heading
% do not use \section anymore after \appendix, only \section*
% is possibly needed

% use appendices with more than one appendix
% then use \section to start each appendix
% you must declare a \section before using any
% \subsection or using \label (\appendices by itself
% starts a section numbered zero.)
%

\ifCLASSOPTIONcaptionsoff
  \newpage
\fi

\bibliographystyle{IEEEtran}
\bibliography{IEEEabrv,ref}

% trigger a \newpage just before the given reference
% number - used to balance the columns on the last page
% adjust value as needed - may need to be readjusted if
% the document is modified later
%\IEEEtriggeratref{8}
% The "triggered" command can be changed if desired:
%\IEEEtriggercmd{\enlargethispage{-5in}}

% references section

% can use a bibliography generated by BibTeX as a .bbl file
% BibTeX documentation can be easily obtained at:
% http://mirror.ctan.org/biblio/bibtex/contrib/doc/
% The IEEEtran BibTeX style support page is at:
% http://www.michaelshell.org/tex/ieeetran/bibtex/
%\bibliographystyle{IEEEtran}
% argument is your BibTeX string definitions and bibliography database(s)
%\bibliography{IEEEabrv,../bib/paper}
%
% <OR> manually copy in the resultant .bbl file
% set second argument of \begin to the number of references
% (used to reserve space for the reference number labels box)

% You can push biographies down or up by placing
% a \vfill before or after them. The appropriate
% use of \vfill depends on what kind of text is
% on the last page and whether or not the columns
% are being equalized.

%\vfill

% Can be used to pull up biographies so that the bottom of the last one
% is flush with the other column.
%\enlargethispage{-5in}

% that's all folks
\end{document}